\documentclass[a4paper]{eccomas_paper-2024}
\usepackage{graphicx}

\usepackage{hyperref}

\usepackage{graphicx}
\usepackage{subcaption}
\usepackage{amsmath}
\usepackage{amsfonts}
\usepackage{amssymb}

\title{TOWARDS ACTIVE FLOW CONTROL STRATEGIES THROUGH DEEP REINFORCEMENT LEARNING}

\author{R. MONTALÀ$^1$, B. FONT$^2$, P. SUÁREZ$^3$, J. RABAULT$^4$, O. LEHMKUHL$^5$, R. VINUESA$^3$ AND I. RODRIGUEZ$^1$}

\heading{R. Montalà, B. Font, P. Suárez, J. Rabault, O. Lehmkuhl, R. Vinuesa and I. Rodriguez}

\address{$^{1}$ Turbulence and Aerodynamics Research Group (TUAREG)\\
Universitat Politècnica de Catalunya (UPC)\\
Terrassa, Spain
\and
$^{2}$ Faculty of Mechanical Engineering\\
Delft University of Technology (TU Delft)\\
Delft, Netherlands
\and 
$^{3}$ FLOW, Engineering Mechanics\\
KTH Royal Institute of Technology\\
Stockholm, Sweden
\and
$^{4}$ Independent Researcher\\
Oslo, Norway
\and
$^{5}$ Computer Applications in Science and Engineering (CASE)\\
Barcelona Supercomputing Center (BSC)\\
Barcelona, Spain
}


\abstract{This paper presents a deep reinforcement learning (DRL) framework for active flow control (AFC) to reduce drag in aerodynamic bodies. Tested on a 3D cylinder at $Re=100$, the DRL approach achieved a $9.32\%$ drag reduction and a $78.4\%$ decrease in lift oscillations by learning advanced actuation strategies. The methodology integrates a CFD solver with a DRL model using an in-memory database for efficient communication between the two instances, making it scalable to more complex flows and higher Reynolds numbers.}

\begin{document}
\thispagestyle{empty}

\section{INTRODUCTION}
In light of the current climate crisis, the transportation industry faces significant challenges in reducing fossil fuel emissions to mitigate the adverse effects of climate change. In particular, the aviation sector already accounts for about $3$\% of the global CO$_2$ emissions \cite{Liu2023}. Reducing these emissions may be possible by exploring innovative methods to decrease drag. In this regard, active flow control (AFC) has demonstrated promising results in controlling the flow around wings \cite{Rodriguez2020,Atzori2021}. However, conventional AFC methods that rely on fixed actuation laws are inherently limited as they can only target specific frequencies within the full spectrum of turbulence scales, leading to a maximum level of drag reduction. Moreover, their fixed-loop nature prevents them from adjusting to instantaneous flow conditions, restricting their applications to dynamical systems that are not continuously evolving. Tunning the actuation parameters can also be challenging and may require extensive trial-and-error, especially in highly turbulent flows.

The emergence of machine learning (ML), coupled with advances in computational power, has revolutionized the state of the art in scientific computing. In the context of AFC, the use of deep reinforcement learning (DRL) appears particularly well-suited for discovering more complex actuation strategies and has already shown its potential in the field of flow control \cite{Garnier2021}. As far as the author is concerned, the first successful application of DRL to AFC was the work by Rabault et al. \cite{Rabault2019}, who considered a two-dimensional cylinder at a Reynolds number of $Re=100$ \cite{Rabault2019}, achieving a drag reduction of approximately $8$\%. The Reynolds number is defined using the fluid density $\rho$, the inflow velocity $U_\infty$, the cylinder diameter $D$ and the fluid dynamic viscosity $\mu$ as $Re=\rho \, U_\infty \, D/\mu$. Based on their previous results, Rabault and Kuhnle \cite{Rabault2019_b} extended their study by implementing a multi-environment approach to explore the capabilities of parallelization in DRL and hence accelerate the training, making the application of DRL affordable for more sophisticated fluid mechanics problems. Building on this data collection parallelization approach, subsequent studies extended the Reynolds number towards higher values. This is the case of Tang et al. \cite{Tang2020}, who investigated the regimes of $Re=100$, $200$, $300$, and $400$ with drag reductions of $5.7$\%, $21.6$\%, $32.7$\%, and $38.7$\%, respectively; and also Varela et al. \cite{Varela2022}, who extended the Reynolds number up to $Re=2,000$ and achieved a reduction of $17.7$\%. The latter authors further applied their DRL set-up to control the wake of a three-dimensional cylinder for the first time, targeting $Re=100$, $200$, $300$ and $400$, and reducing the drag up to $8.0$\%, $17.2$\%, $15.3$\% and $15.1$\%, respectively \cite{Suarez2023}. Other research efforts also combined DRL and AFC to reduce skin friction in wall-bounded flows at $Re_\tau = 180$ \cite{Guastioni2023} or to control the two-dimensional Rayleigh–Bénard convection \cite{Vignon2023}. However, all these studies were confined to canonical problems at low Reynolds numbers, indicating that this methodology is still in its early stages of development.

ML libraries are generally implemented in high-level programming languages, such as Python, while high-performance physics solvers typically rely on low-level languages like C++ or Fortran. This gives rise to the "two-language" problem, where efficiently linking ML models with physical simulations becomes a challenge. In DRL applications for fluid dynamics, which require solving a huge amount of degrees of freedom, the most time-consuming part is the collection of experience data. Therefore, using a fast solver that leverages GPU accelerators, now present in modern HPC clusters, is essential for tackling complex fluid environments, such as those involving intricate geometries or high Reynolds numbers. This study presents an AFC-DRL framework that effectively addresses the "two-language" problem with minimal overhead and employs a GPU-enabled code for computational fluid dynamics (CFD) simulations. This approach ensures rapid trajectory collection for DRL training, making the framework feasible for addressing more complex problems. This framework was first implemented and tested to mitigate the separation bubble in a boundary layer at $Re_\tau=180$ \cite{Font2024}. In the present work, we replicate the case conducted by Suárez et al. \cite{Suarez2023}, extending the validation of the framework to a three-dimensional cylinder at $Re=100$ and evaluating its capabilities in controlling flows around bluff bodies. This represents a significant step towards applying DRL to more realistic industrial scenarios.

\section{METHODOLOGY}

\subsection{DRL set-up}

\begin{figure}[t]
    \centering
    \includegraphics[width=0.9\textwidth]{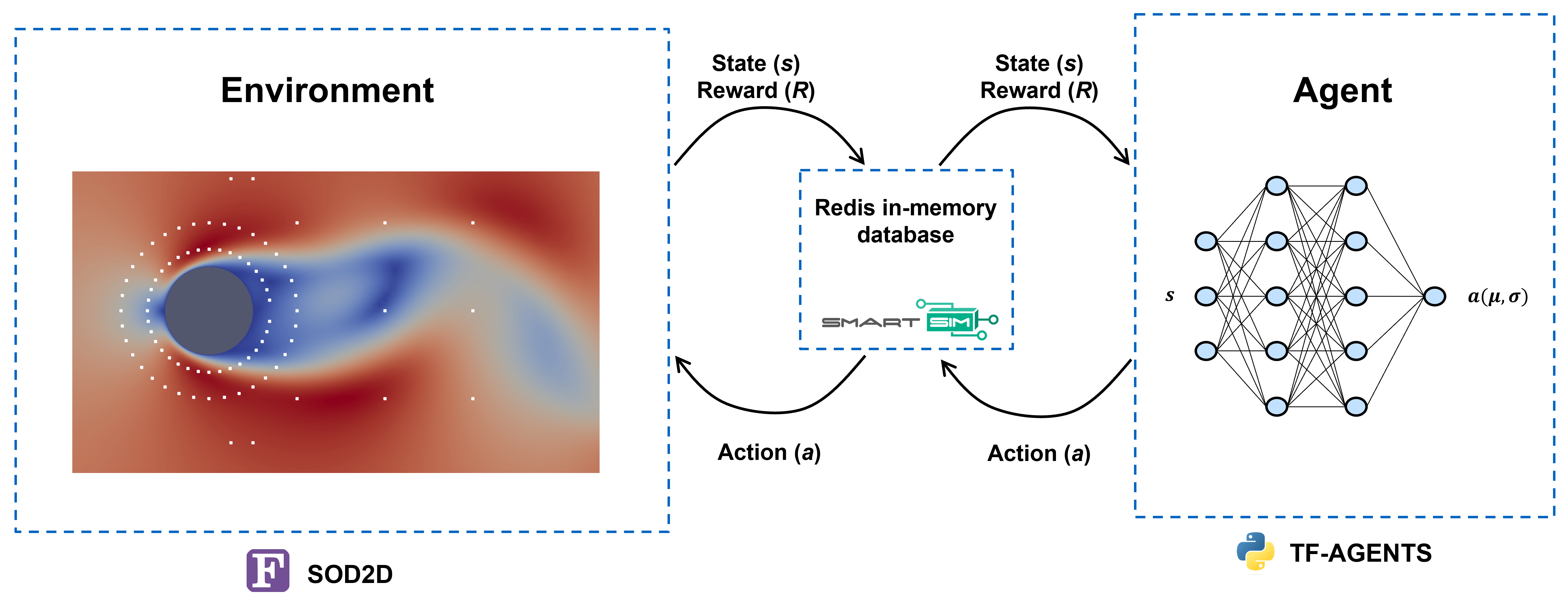}
    \caption{DRL-CFD setup}\label{fig:DRL}
\end{figure}

In DRL, two main entities can be identified: (i) The environment and (ii) the agent. In the current framework, the environment is the CFD simulation that predicts how the flow evolves with a given actuation and the agent is a neuronal network (NN) that predicts the probability distribution of possible action given the state of the environment.

For the environment, the CFD solver called SOD2D \cite{Gasparino2024} is employed. This is a Spectral Element Method (SEM) and GPU-enabled code developed at the Barcelona Supercomputing Center (BSC). The incompressible Navier-Stokes equations are solved, as shown in Eq. \ref{eq:NS_1} and \ref{eq:NS_2}.

\vskip-.6cm
\begin{eqnarray}
\label{eq:NS_1} \nabla \cdot \mathbf{u} = 0 \\
\label{eq:NS_2} \frac{\partial \mathbf{u}}{\partial t} + (\mathbf{u} \cdot \nabla)\mathbf{u} -  \nu \nabla^2 \mathbf{u} + \nabla p = 0
\end{eqnarray}

On the other hand, the agent is built using the Python library TF-Agents \cite{Guadarrama2018}. To solve the so-called "two-language" problem, a Redis in-memory database is used, which is managed through the library SmartSim \cite{Partee2022}, allowing the communication between the CFD model (Fortran) and the DRL agent (Python) with minimal overhead. This workflow was initially proposed in the Relexi project \cite{Kurz2022,Kurz2023}, and adapted to link with the SOD2D CFD solver in \cite{Font2024}. The framework is schematized in Figure \ref{fig:DRL}.

The whole idea of the DRL is that the agent receives the state, e.g., some probes located in the domain, and this returns an action that will be applied back into the environment, e.g., the mass flow rate of the jet. However, to correctly decide the best actuation, the DRL agent needs to be previously trained. To do so, during the training, the agent also receives a reward, representing the magnitude that aims to be optimized. Then, the environment and the agent engage in a trial-and-error process structured as episodes. An episode, in this context, denotes a simulation period wherein the CFD solver and the DRL agent exchange information, including states, actions, and rewards. Following the completion of an episode, this data is used to refine the agent through a training step. The proximal policy optimization (PPO) algorithm \cite{Schulman2017} is used in this case.

\subsection{Case Configuration}
The flow past a cylinder at $Re=100$ is simulated using SOD2D. The three-dimensional domain extends $L_x=30$, $L_y=15D$ and $L_z=4D$ in the streamwise, crosswise and spanwise directions, respectively. The cylinder is located approximately in the middle of the domain, i.e., at $(x,y)=(7.5D, 7.5D)$, and is infinitely long along the spanwise direction. Hence, periodic boundary conditions are applied in this direction. At the inlet, a constant freestream velocity $U_\infty$ is imposed. The slip condition is enforced at the top and bottom surfaces, while zero-gradient conditions with constant pressure are applied at the outlet.

On the cylinder walls, the no-slip condition is applied. When AFC is activated, two sets of $n_{jet}=10$ actuators are distributed along the spanwise direction of the cylinder. Each set contains two jets, one on the top surface of the cylinder ($\theta_{top} = 90^\circ$) and the other on the bottom surface ($\theta_{bot} = 270^\circ$). The two actuators are forced to have the opposite mass flow rate, i.e. $Q_{top} = -Q_{bot}$, so that the mass is conserved instantaneously. In the xy-plane, the actuators have a width of $\omega = 10^\circ$, while in the spanwise direction they extend a width of $L_{jet}=0.4D$. 

\begin{figure}[t]
    \centering
    \includegraphics[width=0.9\textwidth]{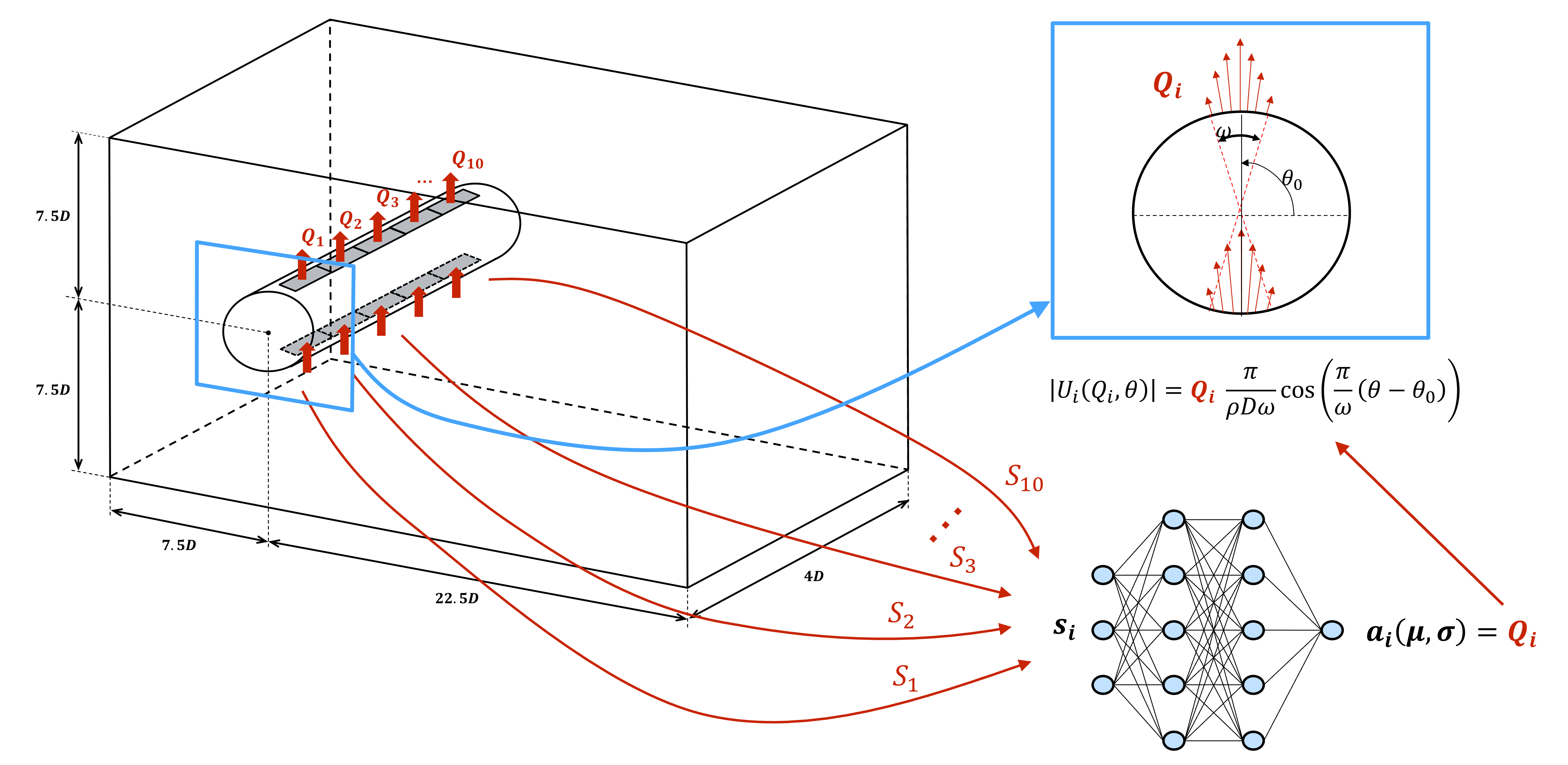}
    \caption{Case configuration}\label{fig:Case}
\end{figure}

Each set of two actuators represents a pseudo-environment. This is schematized in Figure \ref{fig:Case}. Thus, the whole domain is divided into ten pseudo-environments with a width of $L_{jet}$. Moreover, four different CFD simulations are run in parallel. This allows the DRL agent to collect several experiences in parallel, as was previously done in Varela et al. \cite{Varela2022}, speeding up the training. In total, 40 trajectories (10 pseudo-environments x 4 CFD simulations) are collected after each action (batch size). The training is selected to last 50 episodes. The duration of an episode $T_{episode}$ includes six vortex sheddings and in each episode, a total of 120 actions are applied, i.e., $T_{action} = T_{episode}/120$. For the state, each pseudo-environment contains in its z-middle location a slice of 85 witness points distributed around the cylinder walls. The approximate locations of these witness points are illustrated in Figure \ref{fig:DRL}. However, the size of the NN input layer is 85 x 3 = 255, as it takes also into account the state of the two neighbouring pseudo-environments. The NN consists of two hidden layers, each containing 512 neurons.

As the output, the DRL agent is responsible for predicting the optimal mass flow rate $Q$ to maximize the reward. This is then used to compute the velocity profile that is applied as the boundary condition along the jet surfaces, as described in Eq. \ref{eq:vel_jet}; with the velocity perpendicular to the cylinder wall. The minimum and maximum allowable mass flow rates are set to $Q_{min, max}=[-0.176, 0.176]$.

\vskip-.6cm
\begin{eqnarray}
\label{eq:vel_jet} [u_{jet}, v_{jet}, w_{jet}]= Q \frac{\pi}{\rho D \omega}cos(\frac{\pi}{\omega}(\theta-\theta_0)) [cos \theta, sin \theta, 0]
\end{eqnarray}

The reward to train the model is described in Eq. \ref{eq:reward_1}; the first part rewarding the reduction of the drag coefficient $C_d = d/(1/2 \rho U_\infty^2 S)$ with respect to the baseline scenario $C_{d,b}$, and the second penalizing the increase of the lift $C_l = l/(1/2 \rho U_\infty^2 S)$, with $\alpha$ being a weighting factor to adjust the importance of each term. In this case, the reference surface $S$ is defined in terms of the cylinder diameter $D$ and the spanwise length of the jet (or pseudo-environment) $L_{jet}$ as $S = L_{jet} D$, and the drag $d$ and lift $l$ forces correspond to the resulting components of the aerodynamic force in the streamwise and cross-stream directions relative to the freestream $U_\infty$. Finally, the reward applied to the model is computed accounting for the rewards obtained in the other pseudo-environments as shown in Eq. \ref{eq:reward_2}, with $\beta$ being a weighting factor to adjust the importance of the local reward $r_i$ versus the mean rewards of all pseudo-environments $\sum_{j=1}^{n_{jets}} r_j$. The weighting factors in Eq. \ref{eq:reward_1} and Eq. \ref{eq:reward_2} are set to $\alpha=0.3$ and $\beta=0.8$, respectively.

\vskip-.6cm
\begin{eqnarray}
\label{eq:reward_1} r_i = (C_{d,b} - C_d) - \alpha |C_l| \\
\label{eq:reward_2} R_i = \beta r_i + (1-\beta)/n_{jets}\sum_{j=1}^{n_{jets}} r_j
\end{eqnarray}

Note that, to validate the results and compare with previous studies, the set-up presented here mainly mimics the configuration used by Suárez et al. \cite{Suarez2023}.

\section{RESULTS AND DISCUSSION}
\subsection{Training mode}

\begin{figure}[h!]
    \centering
    \includegraphics[width=0.7\textwidth]{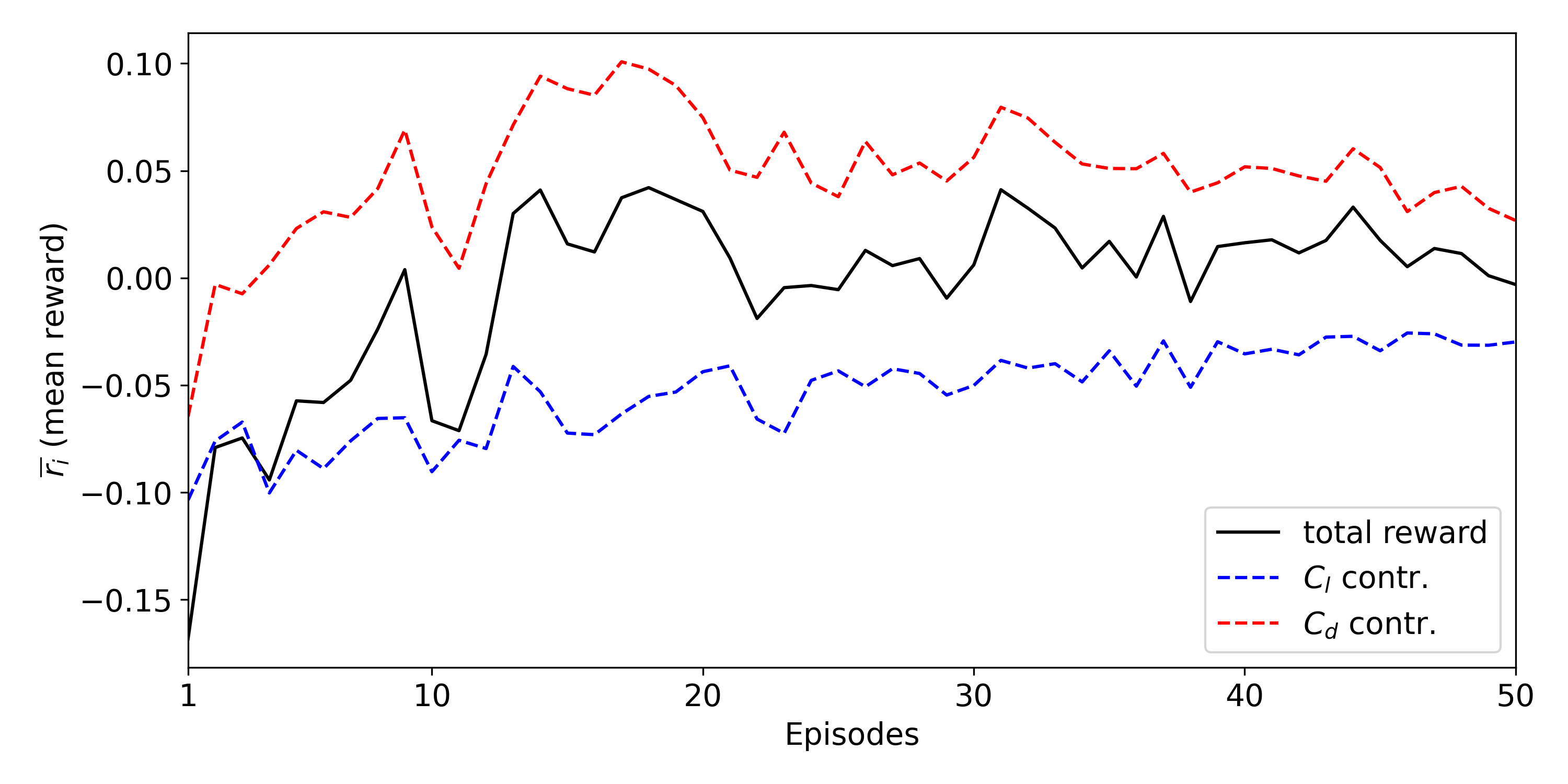}
    \caption{Evolution of the averaged reward $\overline{r_i}$ across the 40 pseudo-environments during the training}\label{fig:reward}
\end{figure}

During the training, the DRL agent adjusts the weights and biases of the NN to maximize the reward described in Eq. \ref{eq:reward_1} and Eq. \ref{eq:reward_2}. Thus, as can be observed from these equations, the DRL control aims at reducing the drag coefficient and also reducing the lift oscillations. The curves showing how the reward evolves along the different training episodes are depicted in Figure \ref{fig:reward}, where the contribution of the lift and drag terms are plotted separately as well. This figure shows the mean reward $\overline{r_i}$ described in Eq. \ref{eq:reward_1} among the forty pseudo-environments. It can be observed that the agent successfully learns a strategy that reduces the drag coefficient and also the oscillations of the lift coefficient. At the beginning of the training, a considerable drag reduction is achieved. As the episodes continue, the lift contribution to the reward progressively increases, while the drag contribution slightly decreases. Overall, the total reward increases. To adjust which is the final objective of the training, e.g., only minimize the drag, the $\alpha$ factor in Eq. \ref{eq:reward_1} could be tuned. To evaluate the learnt actuation law, the model has to be run in deterministic mode and hence apply the learnt policy without any further exploration.

\subsection{Deterministic mode}

\begin{figure}[h!]
    \centering
    \begin{subfigure}[b]{0.45\textwidth}
        \centering
        \includegraphics[width=\textwidth]{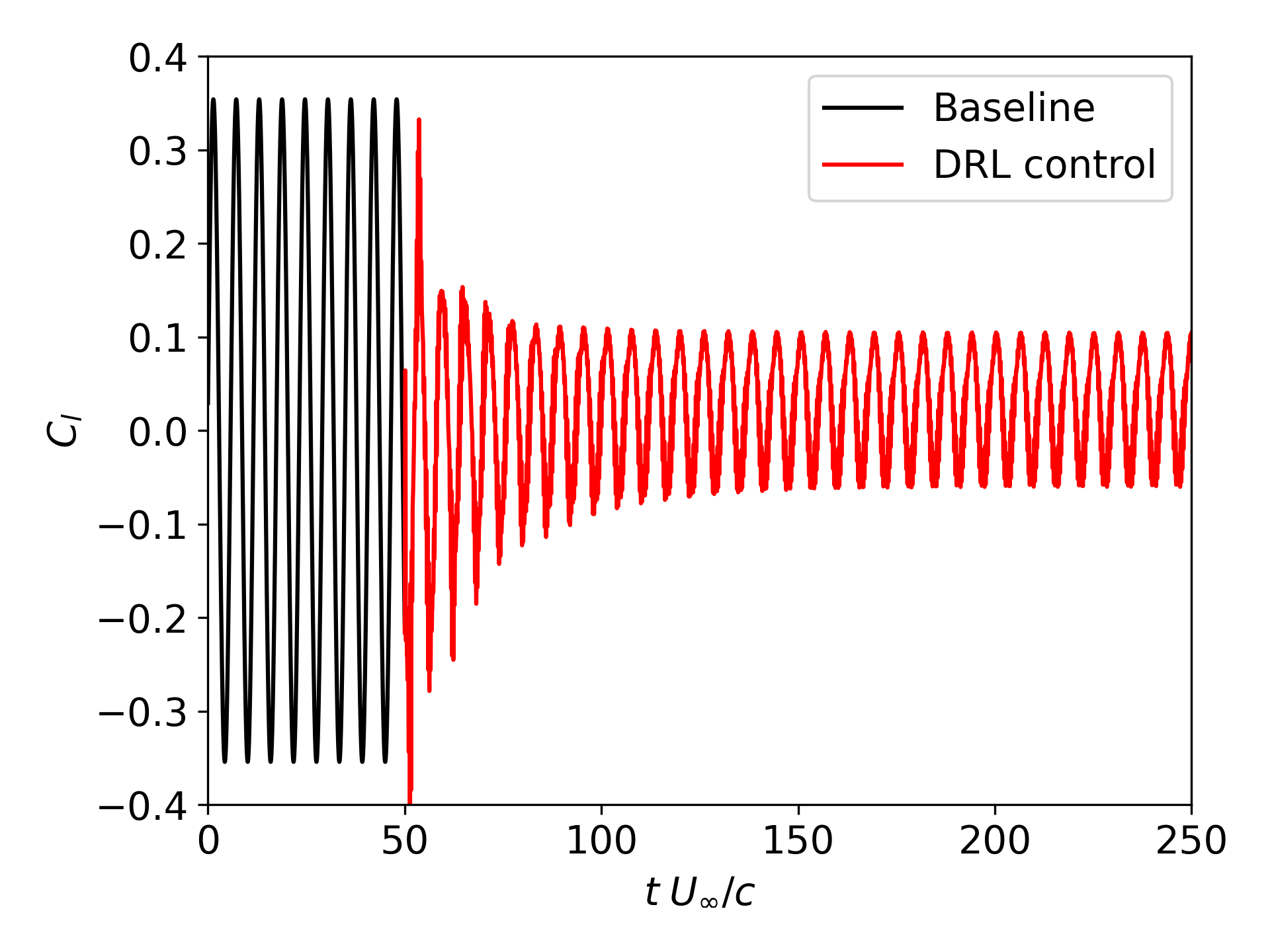}
        \caption{}\label{fig:ClCd_a}
    \end{subfigure}
    \begin{subfigure}[b]{0.45\textwidth}
        \centering
        \includegraphics[width=\textwidth]{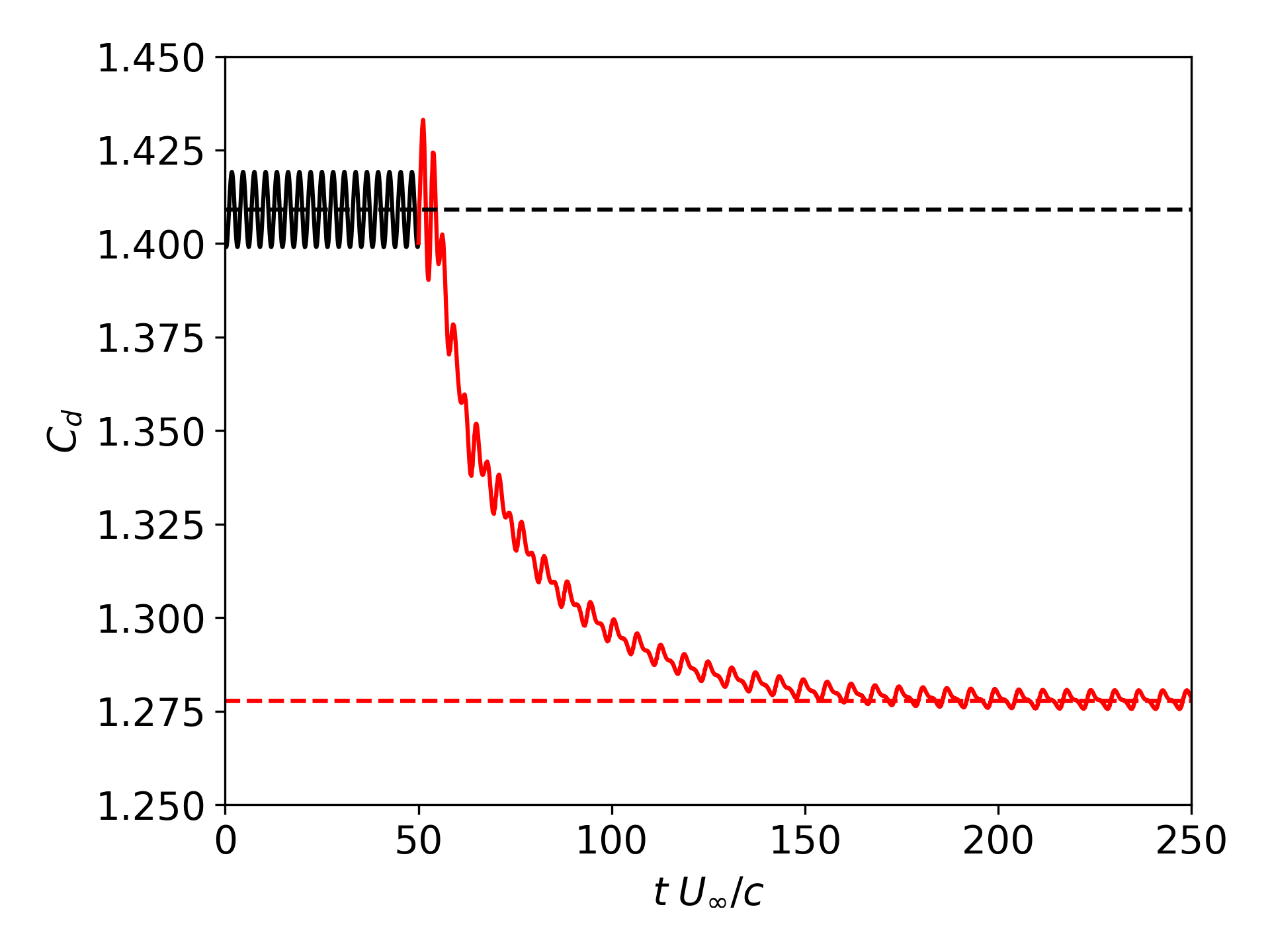}
        \caption{}\label{fig:ClCd_b}
    \end{subfigure}
    \caption{Lift $C_l$ (a) and drag $C_d$ (b) coefficients before and after the DRL control is applied}\label{fig:ClCd}
\end{figure}

Once the DRL agent is run in deterministic mode, the obtained lift and drag coefficients are shown in Figure \ref{fig:ClCd}. Note that the actuation is activated at $tU_\infty/D = 50$, which allows to compare the baseline case ($tU_\infty/D < 50$) against the DRL control ($tU_\infty/D > 50$). After the actuation is applied, the drag coefficient starts to decrease considerably, as well as the amplitude of the lift coefficient. The simulation is run until the control converges into a periodic behaviour. In Table \ref{tab:statistics}, the mean lift and drag coefficients, with the pressure and viscous contributions, are reported. This table also shows the Strouhal number $St_{C_l}$ and the standard deviation $\sigma_{C_l}$ of the lift coefficient signal.  The results indicate that the DRL control achieves a $9.32\%$ reduction in the drag coefficient and a $78.4\%$ decrease in the standard deviation of the lift coefficient. This drag reduction is in close agreement with the $9.4\%$ reduction reported by Suárez et al. \cite{Suarez2023}. It is important to note that all statistics reported here are computed after the control has reached a statistically steady state.

\begin{table}[h!]
    \caption{Lift and drag coefficients statistics}\label{tab:statistics}
    \begin{center}
    \begin{tabular}{*{7}{c}}
    \hline
     & $C_l$ & $\sigma_{C_l}$ & $St_{C_l}$ & $C_d$ & $C_{d,press.}$ & $C_{d,visc.}$  \\
    \hline
    Baseline & 0.012 & 0.250 & 0.170 & 1.409 & 1.051 & 0.358\\
    DRL control & 0.029 & 0.054 & 0.161 & 1.278 & 0.946 & 0.331\\
    \hline
    \end{tabular}
    \end{center}
\end{table}

\begin{figure}[h!]
    \centering
    \includegraphics[width=0.7\textwidth]{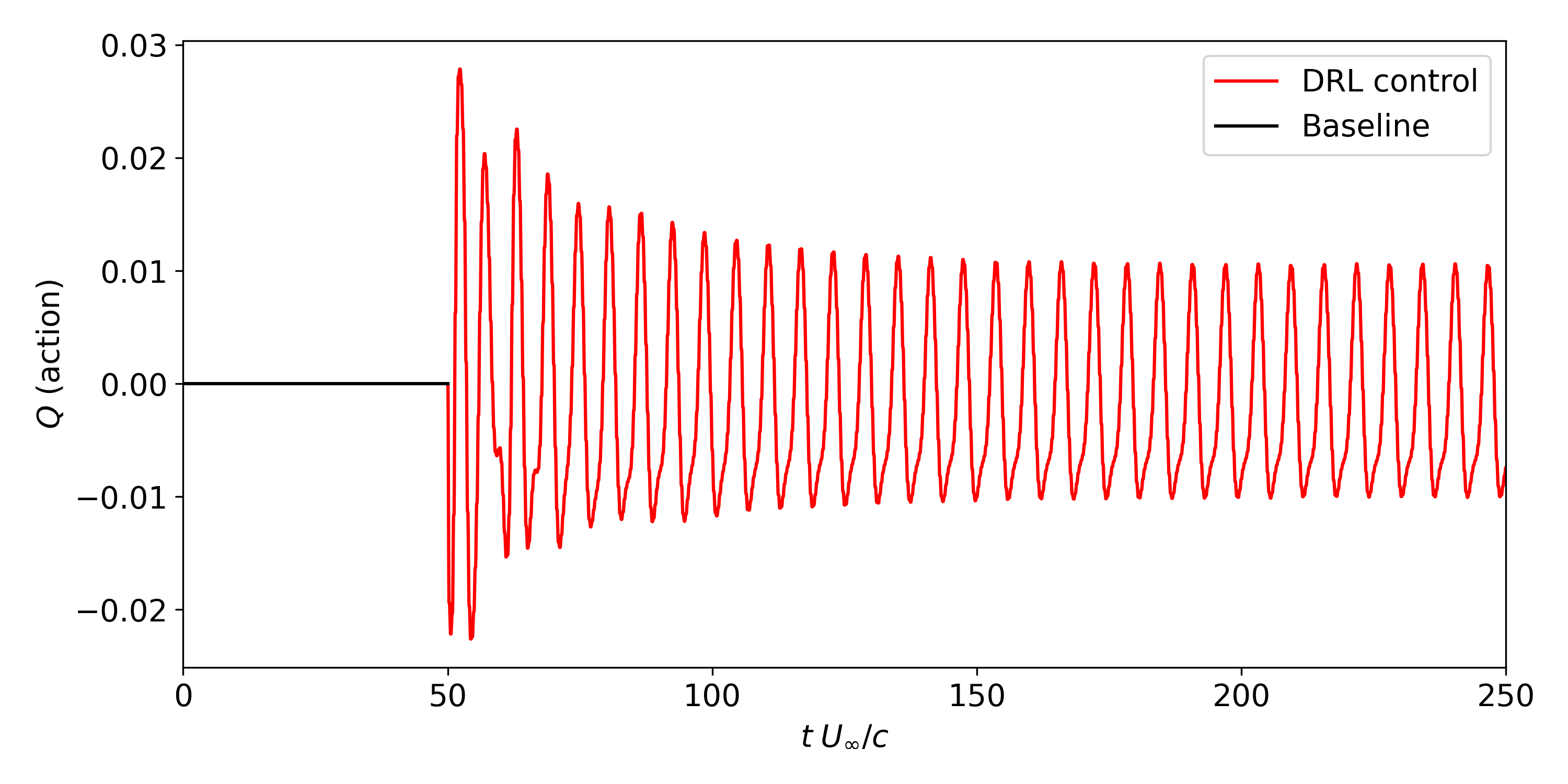}
    \caption{Applied mass flow rate $Q$}\label{fig:action}
\end{figure}

The applied actuation by the DRL agent can be visualized in Figure \ref{fig:action}. After the initial transient, the DRL actuation converges to a periodic signal with a frequency of $St=0.161$. This is approximately the $95\%$ of the baseline vortex-shedding frequency, delaying the vortex-shedding of the controlled scenario (see $St_{C_l}$ in Table \ref{tab:statistics}). The minimum and maximum actions applied by the DRL agent are $Q \in [-0.01, 0.01]$. Nevertheless, the predicted action is not a perfect sinusoidal signal and, at the beginning of each period, a little hump is detected. This is also manifested in the spectral domain, where a second harmonic at $St=0.320$ is obtained.  Compared to a classical periodic control, where a simple sinusoidal actuation would be considered, in this case the agent has learned a double-lobed signal, hence demonstrating that DRL can help to learn more complex control actions that allow to push the limits of drag reduction using AFC. It is worth pointing out that, despite being a three-dimensional domain, the wake of the cylinder at this low $Re$ is essentially two-dimensional. Thus, the applied $Q$ in all the pseudo-domains is exactly the same, as well as the obtained $C_l$ and $C_d$. 

\begin{figure}[h!]
    \centering
    \begin{subfigure}[b]{0.7\textwidth}
        \centering
        \includegraphics[width=\textwidth]{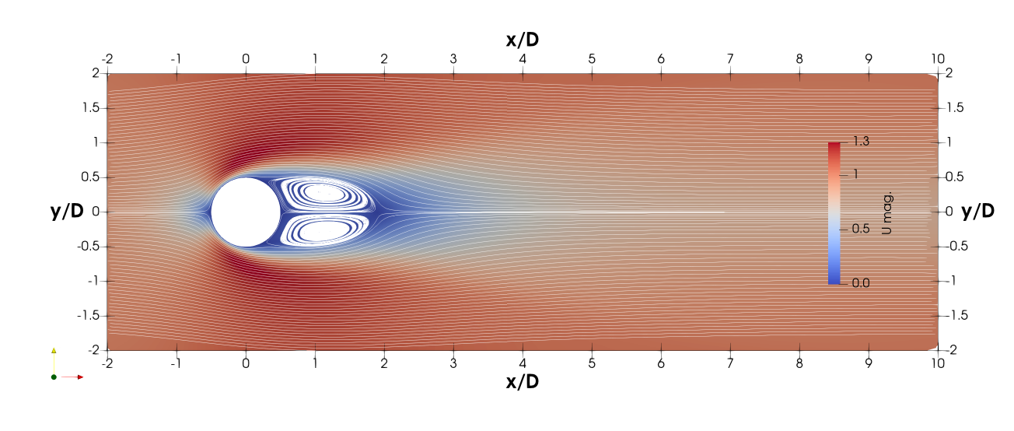}
        \caption{}\label{fig:stream_a}
    \end{subfigure}
    \begin{subfigure}[b]{0.7\textwidth}
        \centering
        \includegraphics[width=\textwidth]{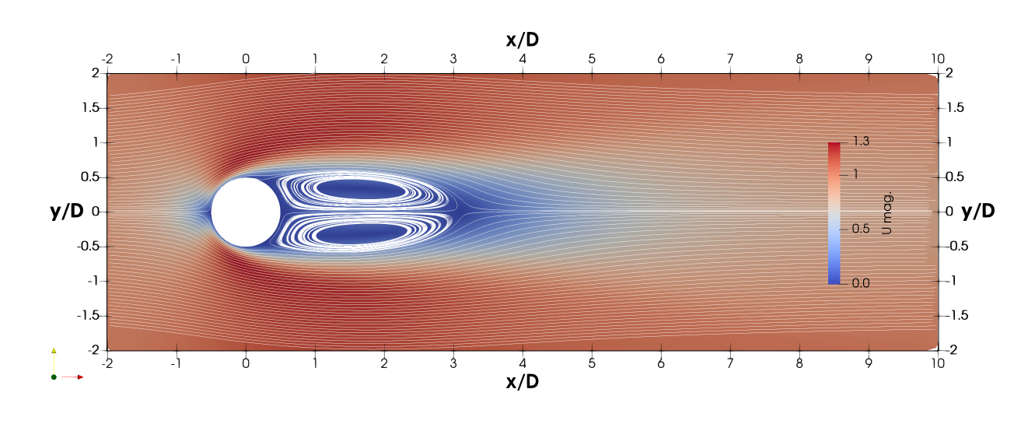}
        \caption{}\label{fig:stream_b}
    \end{subfigure}
    \caption{Velocity streamlines of the baseline (a) and controlled (b) cases}\label{fig:stream}
\end{figure}

\begin{figure}[h!]
    \centering
    \includegraphics[width=0.5\textwidth]{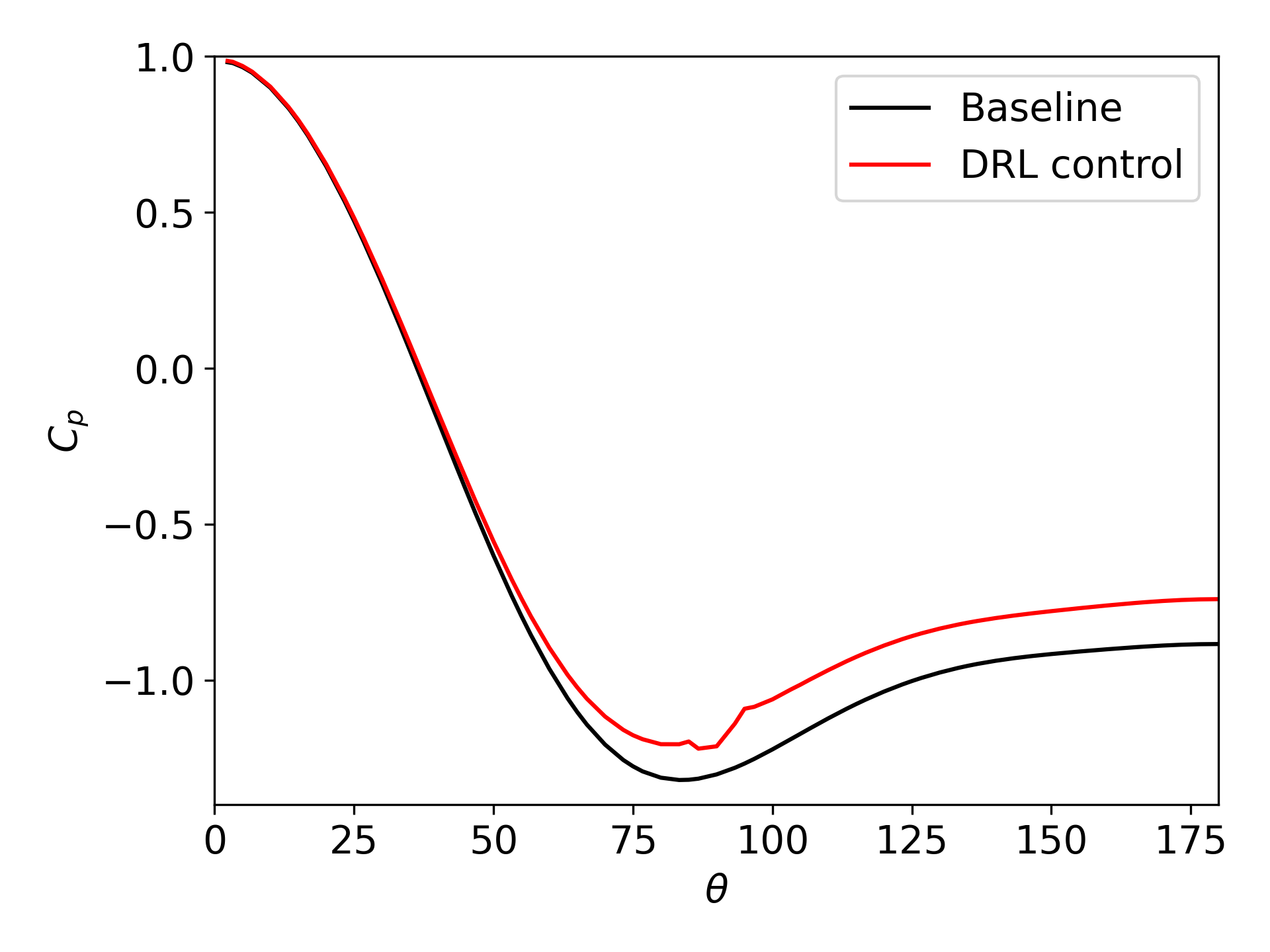}
    \caption{Pressure coefficient $C_p$ along the cylinder wall}\label{fig:Cp}
\end{figure}

As observed in Figure \ref{fig:stream}, the DRL control increases the streamwise length of the wake recirculation bubble, reducing the intensity of the shear layers developed on the cylinder top and bottom surfaces and hence reducing the vortex-shedding frequencies. As depicted in Figure \ref{fig:Cp}, this increases the pressure coefficient $C_p$ on the rear part of the cylinder ($\theta>50^\circ$), finally leading to the drag and lift reductions described before.

\section{CONCLUSIONS}
This study extends the application of the DRL framework initially implemented in Font et al. \cite{Font2024}, where an in-memory database enables the linking of the CFD solver SOD2D with a Python-based DRL model with minimal overhead. In this work, we apply a DRL control strategy to the flow around a three-dimensional cylinder at $Re=100$, following the set-up by Suárez et al. \cite{Suarez2023}. This previous work serves as a benchmark to validate the current framework.

According to the results, the methodology demonstrates its capability to control the flow around a bluff body. The cylinder drag is reduced by approximately $9.32\%$ and the lift oscillations by $78.4\%$. This is consistent with the drag reduction reported by Suárez et al. \cite{Suarez2023}. This improvement is accomplished by controlling the intensity of the shear layers formed around the cylinder top and bottom regions, thereby delaying the onset of instabilities. 

Despite the relative simplicity of this case, the results highlight the potential of DRL to find more advanced actuation strategies compared to traditional periodic controls. Additionally, integrating the DRL model with the GPU-accelerated SOD2D CFD code allows for the rapid collection of a larger volume of experiences, making the framework scalable and feasible for more challenging cases involving higher Reynolds numbers and/or complex geometries.

\section*{Acknowledgements}

This research has received financial support from the {\it Ministerio de Ciencia e Innovación} of Spain (PID2020-116937RB-C21 and PID2020-116937RB-C22). Simulations were conducted with the assistance of the {\it Red Española de Supercomputación}, who granted computational resources to the HPC facility of {\it MareNostrum 5}, at the Barcelona Supercomputing Center (IM-2024-2-0004). Ricard Montalà work is funded by a FI-SDUR grant (2022 FISDU 00066) from the {\it Departament de Recerca i Universitats de la Generalitat de Catalunya}. Oriol Lehmkuhl has been partially supported by a {\it Ramon y Cajal} postdoctoral contract (RYC2018-025949-I). The authors also acknowledge the support of the {\it Departament de Recerca i Universitats de la Generalitat de Catalunya} through the research group Large-scale Computational Fluid Dynamics (2021 SGR 00902) and the Turbulence and Aerodynamics Research Group (2021 SGR 01051).


\begin{thebibliography}{99}

\bibitem{Liu2023} Liu, Z., Deng, Z., Davis, S. and Ciais, P. "Monitoring global carbon emissions in 2022". \textit{Nat. Rev. Earth Environ.} (2023) \textbf{4}:205--206. https://doi.org/10.1038/s43017-023-00406-z

\bibitem{Rodriguez2020} Rodriguez, I., Lehmkuhl, O. and Borrell, R. "Effects of the actuation on the boundary layer of an airfoil at Reynolds number Re = 60,000". \textit{Flow Turbul. Combust.} (2020) \textbf{105}:607--626. https://doi.org/10.1007/s10494-020-00160-y

\bibitem{Atzori2021} Atzori, M., Vinuesa, R., Stroh, A., Gatti, D., Frohnapfel, B. and Schlatter, P. "Uniform blowing and suction applied to nonuniform adverse-pressure-gradient wing boundary layers". \textit{Phys. Rev. Fluids} (2021) \textbf{6}:113904. https://doi.org/10.1103/PhysRevFluids.6.113904

\bibitem{Garnier2021} Garnier, P., Viquerat, J., Rabault, J., Larcher, A., Kuhnle, A. and Hachem, E. "A review on deep reinforcement learning for fluid mechanics". \textit{Comput. Fluids} (2021) \textbf{225}:104973. https://doi.org/10.1016/j.compfluid.2021.104973

\bibitem{Rabault2019} Rabault, J., Kuchta, M., Jensen, A., Réglade, U. and Cerardi, N. "Artificial neural networks trained through deep reinforcement learning discover control strategies for active flow control". \textit{J. Fluids Mech.} (2019) \textbf{865}:281--302. https://doi.org/10.1017/jfm.2019.62

\bibitem{Rabault2019_b} Rabault, J. and Kuhnle, A. "Accelerating deep reinforcement learning strategies of flow control through a multi-environment approach". \textit{Phys. Fluids} (2019) \textbf{31}:094105. https://doi.org/10.1063/1.5116415

\bibitem{Tang2020} Tang, H., Rabault, J., Kunhle, A., Wang, Y. and Wang, T. "Robust active flow control over a range of Reynolds numbers using an artificial neural network trained through deep reinforcement learning". \textit{Phys. Fluids} (2020) \textbf{32}:053605. https://doi.org/10.1063/5.0006492

\bibitem{Varela2022} Varela, P., Suárez, P., Alcántara-Ávila, F., Miró, A., Rabault, J., Font, B., García-Cuevas, L. M.,  Lehmkuhl, O. and Vinuesa, R. "Deep reinforcement learning for flow control exploits
different physics for increasing Reynolds number regimes". \textit{Actuators} (2022) \textbf{11}:359. https://doi.org/10.3390/act11120359

\bibitem{Suarez2023} Suárez, P., Alcántara-Ávila, F., Miró, A., Rabault, J., Font, B.,  Lehmkuhl, O. and Vinuesa, R. "Active flow control for three-dimensional cylinders through deep reinforcement learning". In: \textit{14th International ERCOFTAC Symposium on Engineering, Turbulence, Modelling and Measurements} (2023)

\bibitem{Guastioni2023} Guastioni, L., Rabault, J., Schlatter, P., Azizpour, H. and Vinuesa, R. "Deep reinforcement learning for turbulent drag reduction in channel flows". \textit{Eur. Phys. J. E} (2023) \textbf{46}. https://doi.org/10.1140/epje/s10189-023-00285-8

\bibitem{Font2024} Font, B., Alcántara-Ávila, F., Rabault, J., Vinuesa, R. and Lehmkuhl, O. "Active flow control of a turbulent separation bubble through deep reinforcement learning". \textit{J. Phys. Conf. Ser.} (2024) \textbf{2753}: 012022. https://doi.org/10.1088/1742-6596/2753/1/012022

\bibitem{Vignon2023} Vignon, C., Rabault, J., Vasanth, J., Alcántara-Ávila, F., Mortensen, M. and Vinuesa, R. "Effective control of two-dimensional Rayleigh–Bénard convection: Invariant multi-agent reinforcement learning is
all you need". \textit{Phys. Fluids} (2023) \textbf{35}:065146. https://doi.org/10.1063/5.0153181

\bibitem{Gasparino2024} Gasparino, L., Spiga, F. and Lehmkuhl, O. "SOD2D: A GPU-enabled spectral finite elements method for compressible scale-resolving simulations". \textit{Comput. Phys. Comun.} (2024) \textbf{297}:109067. https://doi.org/10.1016/j.cpc.2023.109067

\bibitem{Guadarrama2018} Guadarrama, S., Korattikara, A., Ramirez, O., Castro, P., Holly, E., Fishman, S., Wang, K., Gonina, E., Wu, N., Kokiopoulou, E., Sbaiz, L., Smith, J., Bartók, G., Berent, J., Harris, C., Vanhoucke, V. and Brevdo, E. "TF-Agents: A library for Reinforcement Learning in TensorFlow". (2018). https://github.com/tensorflow/agents

\bibitem{Partee2022} Partee, S., Ellis, M., Rigazzi, A., Shao, A. E., Bachman, S., Marques, G. and Robbins, B. "Using machine learning at scale in HPC simulations with SmartSim: An application to ocean climate modeling". \textit{J. Comput. Sci.} (2022) \textbf{62}:101707. https://doi.org/10.5281/zenodo.4682270

\bibitem{Kurz2022} Kurz, M., Offenhäuser, P., Viola, D., Resch, M. and Beck, A. "Relexi - A scalable open source reinforcement learning framework for high-performance computing". \textit{Software Impacts} (2022) \textbf{14}:100422. http://doi.org/10.1016/j.simpa.2022.100422

\bibitem{Kurz2023} Kurz, M., Offenhäuser, P. and Beck, A. "Deep reinforcement learning for turbulence modelling in large eddy simulations". \textit{Int. J. Heat Fluid Flow} (2023) \textbf{99}:109094. https://doi.org/10.1016/j.ijheatfluidflow.2022.109094

\bibitem{Schulman2017} Schulman, F., Wolski, P., Dhariwal, A., Radford, A. and Klimov, O. "Proximal policy optimization algorithms". (2017). https://doi.org/10.48550/arXiv.1707.06347



\end{thebibliography}
\end{document}